\documentclass[10pt,twocolumn,letterpaper]{article}

\usepackage{cvpr}              

\usepackage{enumerate}
\usepackage{enumitem}
\usepackage{times}
\usepackage{epsfig}
\usepackage{graphicx}
\usepackage{amsmath}
\usepackage{amssymb}
\usepackage{dblfloatfix}
\usepackage{transparent}
\usepackage{float}
\usepackage{subcaption}
\usepackage{sidecap}
\usepackage{booktabs}
\usepackage{tabulary}
\usepackage{multirow}
\usepackage[dvipsnames]{xcolor}
\usepackage[british,english,american]{babel}
\usepackage{soul}
\usepackage{xspace}\xspace
\usepackage{adjustbox}
\usepackage{array}
\usepackage{algorithm}
\usepackage{algorithmic}
\usepackage{listings}
\usepackage{multirow}
\DeclareMathOperator*{\argmax}{argmax}

\usepackage{algorithm}
\usepackage{algorithmic}

\usepackage[pagebackref,breaklinks,colorlinks]{hyperref}

\usepackage[capitalize]{cleveref}
\crefname{section}{Sec.}{Secs.}
\Crefname{section}{Section}{Sections}
\Crefname{table}{Table}{Tables}
\crefname{table}{Tab.}{Tabs.}

\def\cvprPaperID{11} 
\def\confName{CVPR}
\def\confYear{2022}

\begin{document}

\title{Consistency-based Active Learning for Object Detection}

\author{Weiping Yu$^{1}$, Sijie Zhu$^2$, Taojiannan Yang$^{2}$, Chen Chen$^{2}$\\
$^1$School of Computer Science and Engineering, Nanyang Technological University, Singapore\\
$^2$Center for Research in Computer Vision, University of Central Florida, USA\\
{\tt\small weiping001@e.ntu.edu.sg}\\
{\tt\small \{sizhu,taoyang1122\}@knights.ucf.edu; chen.chen@crcv.ucf.edu}
}
\maketitle

\begin{abstract}
Active learning aims to improve the performance of the task model by selecting the most informative samples with a limited budget. Unlike most recent works that focus on applying active learning for image classification, we propose an effective Consistency-based Active Learning method for object Detection (CALD), which fully explores the consistency between the original and augmented data. 
CALD has three appealing benefits. (i) CALD is systematically designed by investigating the weaknesses of existing active learning methods, which do not take the unique challenges of object detection into account. (ii) CALD unifies box regression and classification with a single metric, which is not concerned with active learning methods for classification. CALD also focuses on the most informative local region rather than the whole image, which is beneficial for object detection. (iii) CALD not only gauges individual information for sample selection but also leverages mutual information to encourage a balanced data distribution. Extensive experiments show that CALD significantly outperforms existing state-of-the-art task-agnostic and detection-specific active learning methods on general object detection datasets. Based on the Faster R-CNN detector, CALD consistently surpasses the baseline method (random selection) by 2.9/2.8/0.8 mAP on average on PASCAL VOC 2007, PASCAL VOC 2012, and MS COCO. \textbf{Code is available at} \url{https://github.com/we1pingyu/CALD}

\end{abstract}

\section{Introduction}

One of the biggest bottlenecks of learning-based computer vision algorithms is the scale of annotated data. Recently, various learning methods, e.g. semi-supervised learning~\cite{mahajan2018exploring,berthelot2019mixmatch,xie2019unsupervised} and unsupervised learning~\cite{chen2020simple,chen2020improved,he2020momentum}, exploit information from unlabeled data to avoid the expensive cost of annotating data. Given a fixed labeled data pool, however, semi-supervised learning and unsupervised learning still cannot achieve the upper-bound performance of fully-supervised learning~\cite{mahajan2018exploring}. Active learning~\cite{gal2016dropout,sener2017active,sinha2019variational,zhang2020state,yoo2019learning,gao2020consistency,nguyen2004active,yang2015multi,joshi2009multi,lewis1994heterogeneous,beluch2018power} provides a new perspective for reducing the demand of labeled data by selecting the most informative data (i.e. task models can gain higher performance by training on these samples) to be annotated. It has been proved successful on basic vision tasks, e.g. image classification. 


\begin{table}
\scriptsize

\begin{center}
 \resizebox{\linewidth}{!}{   
\begin{tabular}{l|l|c|c}
\hline

\hline
Methods                & Task           & 1st cycle & 2nd cycle \\
\hline
\multirow{2}{*}{LL4AL~\cite{yoo2019learning}} & classification & +6.5\%     & +10.1\%    \\\cline{2-4} 
                       & detection      & +2.6\%     & +2.7\%     \\
\hline
\multirow{2}{*}{VAAL~\cite{sinha2019variational}}  & classification & +6.2\%     & +6.0\%     \\\cline{2-4} 
                       & detection      & +2.3\%     & +0.0\%       \\
\hline
\textbf{CALD (Ours)}                 & detection      & +8.4\%     & +5.8\%    \\ 
\hline

\hline

\end{tabular}
}
\end{center}
\vspace{-10pt}
\caption{Performance of two classification-based active learning methods (LL4AL~\cite{yoo2019learning} and VAAL~\cite{sinha2019variational}) on classification and detection compared with CALD (Ours) on detection. \ul{The percentage refers to the improvement compared with random selection.} \textbf{The results show that the improvement drops significantly when classification-based active learning methods are transferred from classification to detection}.}
\label{tab:analyse}
\end{table}

As a fundamental and challenging task in computer vision, object detection \cite{carion2020end,yu2021towards,zhu2020deformable,duan2019centernet,li2020density} also suffers from intensive labor and time for data annotation, as it requires both bounding box and class label. Previous works generally follow the spirit of semi-supervised \cite{sohn2020simple,tang2021proposal,NEURIPS2019_d0f4dae8} and unsupervised learning~\cite{xie2021detco,he2020momentum,chen2020improved} to better leverage the unlabeled data, \textit{while not enough effort has been made to improve the efficiency of the annotation process}. 

Although active learning methods \cite{sinha2019variational,yoo2019learning,gao2020consistency,zhang2020state,mayer2020adversarial} are popular for image classification, \textbf{directly applying classification-based active learning methods to object detection does not lead to satisfactory improvement}        (see Table~\ref{tab:analyse}), due to \textbf{three challenges} of this problem. 1) Classification-based methods only consider the predicted class distribution, while the bounding box prediction could be equally important for selecting informative samples in object detection. 2) Informative objects often exist in local regions of images along with other uninformative objects. Simply applying a \textbf{global metric}, e.g. the loss of the model, may ignore some informative objects in an image where most objects are uninformative.
3) There is only one class assigned to each sample for classification, while for object detection, one image could have multiple objects with different classes which may have an unbalanced distribution. As a result, classification-based selection may lead to a highly unbalanced class distribution.

\begin{figure}[t]
\vspace{-3pt}
\begin{center}
 \includegraphics[width=.9\linewidth]{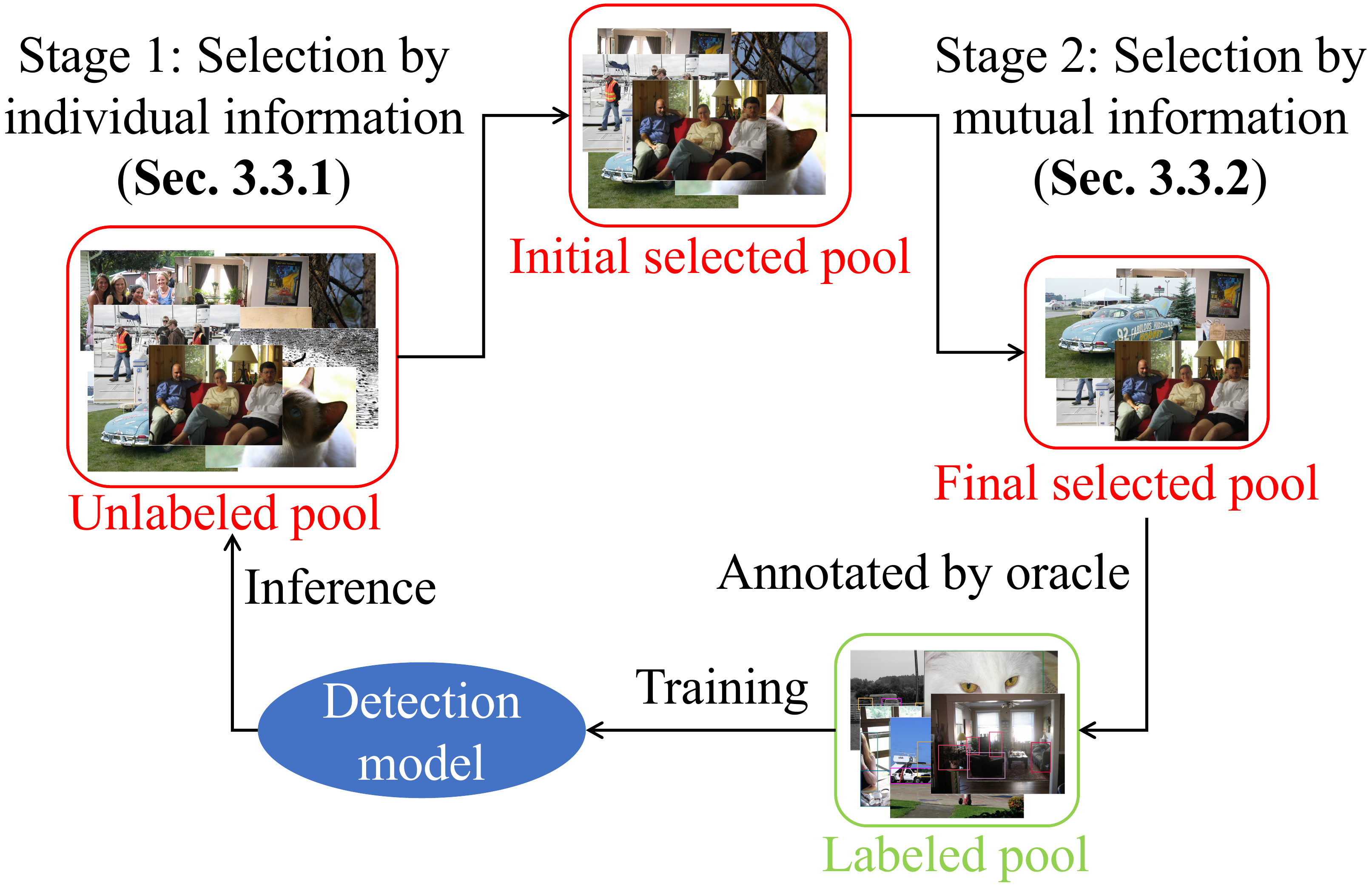}
\end{center}
\vspace{-15pt}
   \caption{A high-level overview of CALD.}
\label{fig:fig1}
\end{figure}

In this paper, we propose a Consistency-based Active Learning framework for object Detection (CALD). As shown in Fig.~\ref{fig:fig1}, in the first stage, we apply data augmentations to each unlabeled image and feed the original and augmented images to the initial detector (trained on randomly selected images before the process of active learning). We then calculate individual information, which unifies box regression and classification and focuses on local regions, based on the consistency between predictions of the original and augmented images to select informative samples to the initial selected pool. In the second stage, we further filter samples in the initial selected pool via mutual information (i.e. distance of class distributions of the selected pool and labeled pool) to \textbf{alleviate unbalanced class distribution}, leading to the final selected pool that meets the annotation budget. The main {contributions} are threefold:
\setlist{nolistsep}
\begin{itemize}[noitemsep,leftmargin=*]
    \item  We discover the gap between active learning for image classification and object detection, which leads to the performances drop when existing classification-based active learning methods are transferred to detection. Our analysis inspires three principled guidelines of how to design an effective detection-specific active learning approach. 
    
    \item We propose a novel detection-specific active learning method, CALD, considering the consistency of both bounding box and predicted class distribution when augmentation is applied to overcome the challenges brought by inconsistencies between classification and detection. 
    
    \item Extensive experiments on PASCAL VOC~\cite{everingham2010pascal}, MS COCO~\cite{lin2014microsoft} with Faster R-CNN~\cite{ren2016faster} and RetinaNet~\cite{lin2017focal} show that CALD outperforms 
    state-of-the-art task-agnostic and detection-specific active learning methods.
\end{itemize}

\section{Related Work}
\label{sec:related}
\noindent \textbf{Label-efficient object detection.} Currently, the most popular and successful object detection frameworks are Convolutional Neural Network (CNN)-based one-stage or two-stage detectors such as Faster R-CNN (FRCNN)~\cite{ren2016faster} and RetinaNet~\cite{lin2017focal}. Compared with image classification~\cite{he2016deep,krizhevsky2012imagenet}, object detectors need to implement both regression and classification tasks on local regions~\cite{hoiem2012diagnosing}. 
From the perspective of annotation, object detection requires not only classification but also bounding box. Various label-efficient methods are proposed to better leverage the information from unlabeled data. Most of them follow a paradigm of unsupervised or semi-supervised learning.

One popular class of semi-supervised learning methods of object detection~\cite{NEURIPS2019_d0f4dae8,sohn2020simple,tang2021proposal,liu2021unbiased} are based on augmentation~\cite{berthelot2019mixmatch,yang2020gradaug,xie2019unsupervised,hendrycks2019augmix,cubuk2019autoaugment} and regularization. The key idea is to first generate pseudo labels for unlabeled images then use them to fine-tune the detector with augmentations.
Another prevailing learning paradigm is unsupervised learning~\cite{chen2020simple,he2020momentum,chen2020improved,xie2021detco} which aims to learn a better representation with unlabeled data. Then the model can be deployed as the backbone for downstream tasks such as object detection.
\emph{All the mentioned methods focus on taking advantage of unlabeled data, while the annotation procedure for supervised training is ignored}.

\noindent \textbf{Classification-based active learning.}
A flurry of active learning methods \cite{sinha2019variational,yoo2019learning,gao2020consistency,zhang2020state,nguyen2004active,krishnamurthy2002algorithms,dagan1995committee,gal2016dropout,beluch2018power,mayer2020adversarial,mahapatra2018efficient} have been proposed for image classification.
The most popular methods are based on pool-based selective sampling~\cite{beluch2018power,yoo2019learning,gal2016dropout,sinha2019variational,sener2017active}.
Pool-based methods continuously select the most informative samples from the unlabeled samples (i.e. unlabeled pool) as selected samples (selected pool) for labeling, and add them to the labeled samples (labeled pool) with a limit of budget. 
Learning Loss for Active Learning (LL4AL)~\cite{yoo2019learning} predicts target losses of unlabeled samples. Higher loss indicates the sample has higher uncertainty under the task model. 



Another representative \textit{task-agnostic} active learning method, called Variational Adversarial Active Learning (VAAL)~\cite{sinha2019variational}, learns a latent space from a VAE and trains an adversarial network to discriminate samples between unlabeled and labeled data.

\noindent \textbf{Detection-specific active learning.}
\ul{Unfortunately, there are limited works using pure active learning for object detection.} Most related works~\cite{desai2019adaptive,haussmann2020scalable,bietti2012active} focus on classification, ignoring box regression or relying on the assistance of semi-supervised learning. The work closely following the standard active learning is 
\cite{kao2018localization}, which introduces two methods: Localization Tightness with the classification information (LT/C) and Localization Stability with the classification information (LS+C). The former is based on the overlapping ratio between the region proposals and the final prediction. Therefore, it can only be applied to two-stage detectors. The latter is based on the variation of predicted object locations when input images are corrupted by noise, which ignores the difference of classification. 

A part of Self-supervised Sample Mining (SSM)~\cite{wang2018towards} can be classified as active learning. SSM takes two steps to select samples: the first step is based on classification and the second step uses copy-paste strategy to cross validate the uncertainty of images. This method can easily lead to a distribution of samples with little diversity.


\begin{figure*}[t]
\begin{center}
 \includegraphics[width=.99\linewidth]{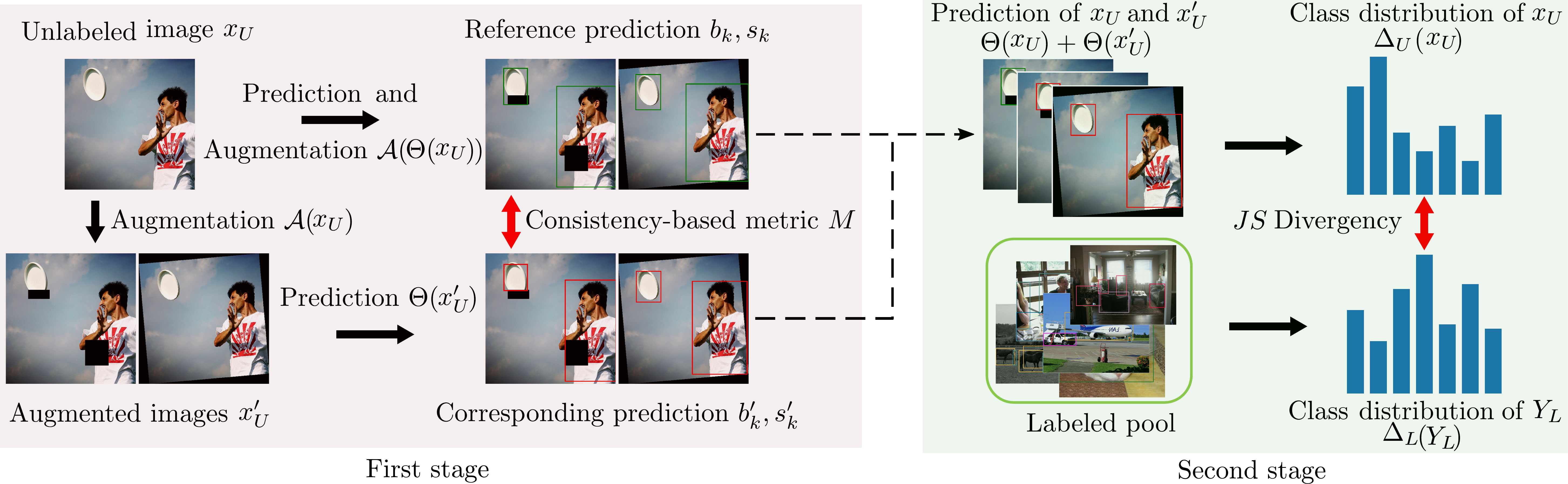}
\end{center}
\vspace{-13pt}
   \caption{The two stages of the proposed CALD. The first stage selects samples based on individual information while the second stage uses mutual information to further refine the selected samples. Individual information is assessed by the consistency-based metric of reference and corresponding predictions. Mutual information refers to the $JS$ divergence of class distributions of an unlabeled image and the labeled pool.}
\label{fig:detail}
\end{figure*}

\section{Method}

\subsection{Problem Overview}
\label{sec:method:problem}
Given a fixed annotation budget, the active learning
paradigm follows an iterative setting with $C$ cycles and each cycle has $1/C$ of the total budget. Each cycle consists of \textbf{metric calculation} (which indicates how much information is in the samples -- the core of active learning for selecting informative samples), data sampling and model training. In the $i$-th cycle, we have unlabeled images $x_{U}$ in the unlabeled pool $X_{U}^{i}$, meanwhile there is also a labeled pool with ground truth $(X_{L}^{i},Y_{L}^{i})$. Active learning aims to select the most informative samples from $X_{U}^{i}$ to the final selected pool $X_{F}^{i}$, annotate them by oracle $\Omega$, and add them to $(X_{L}^{i},Y_{L}^{i})$. The process can be formulated as: 
$(X_{L}^{i+1},Y_{L}^{i+1})=(X_{L}^{i},Y_{L}^{i}) \cup \Omega(X_{F}^{i})$. Since our method has two stages (Fig. \ref{fig:fig1}), we use $X_{I}^{i}$ to denote the initial selected pool of the first stage. We will omit the superscript $i$ since the operation is exactly the same in all the $C$ cycles.

\subsection{Guidelines of A Good Metric -- Motivation}
\label{sec:inconsistency}


Before diving into the details of our proposed method, we first shed light on the following important question as the motivation of our approach.
\textit{The core of active learning lies in finding a good metric that selects the most informative samples, but what is a good metric for object detection? }

The information represented by a reasonable metric should follow two principles: 1) The information of samples (both labeled pool and selected pool) should generally decrease as learning cycle progresses, because the more knowledge the model learns, the less \textbf{new} information that samples can provide. 2) The information of the selected pool using the metric should be higher than that of the labeled pool, because the detectors have already learned most of the information in labeled pool and thus look for samples with more information to improve performance.



The reason why the metric of \textit{predicted loss} cannot properly denote the information in samples when LL4AL~\cite{yoo2019learning} is transferred from classification to detection lies in the inconsistency of the two tasks. There are at least \textbf{three inconsistencies} between them. First, detectors perform both box regression and classification during training and testing, so a single loss cannot represent the two types of uncertainty at the same time. Secondly, detectors focus more on local regions. For example, if LL4AL gives low uncertainty for an image, which means most part of the image is uninformative and some informative patches with small areas may be ignored. Thirdly, since images for the detection task usually contain \textit{multiple objects}, the selected samples may contain some uninformative objects which are more likely to cause unbalanced class distribution than classification.

LS+C~\cite{kao2018localization} partly addresses these issues by computing the average stability of all bounding boxes, while ignoring the classification confidence. It then selects the prediction with the lowest confidence as the most informative patch. The sum of these two metrics is considered as the final metric. However, it still fails to find the most informative combination of box regression and classification.

In contrast to existing methods, \ul{our framework follows three guidelines to tackle the inconsistencies:} 1) Bounding box and classification predictions are considered together in one metric. 2) Our metric is computed based on local regions instead of the average information of the global image. 3) We apply an extra step to \textbf{alleviate the unbalanced class distribution of selected samples}. In the next section, we elaborate the two stages of the proposed CALD.

\subsection{Consistency-based Active Learning for Object Detection (CALD)}
\label{sec:consist}

In each active learning cycle, we have a detector $\Theta$ with parameters $\theta$ and a series of data augmentations $\mathcal{A}$ (e.g. flipping). For an unlabeled image $x_{U}$, its augmented versions are $x'_{U}=\mathcal{A}(x_{U})$. The predictions (i.e. bounding boxes and classification scores) of $x_{U}$ can be represented as $\Theta(x_{U};\theta)$, and they can be mapped onto the augmented images $x'_{U}$. Therefore, we denote the transformation of the $k$-th prediction of $x_U$ as reference prediction including \textbf{reference box} $b_k$, \textbf{reference score} $s_k$,
which can be formulated as:
\begin{equation}
    \{b_k\},\{s_k\}=\mathcal{A}(\Theta(x_{U};\theta))
\end{equation}

Note that class-wise $s_k = [\varphi_{1},\varphi_{2},\cdots,\varphi_{n},\cdots]^\mathrm{T}$, where $\varphi_{n}$ denotes the confidence of the $n$-th class. The way $\mathcal{A}$ transforming predictions is similar to augmentation of images. Take horizontal flipping for instance, we get the box prediction of the augmented image by horizontally flipping the corresponding box from the original image and inheriting the classification prediction.  

The predictions of augmented images obtained directly by the detector $\Theta$ can be expressed as:
\begin{equation}
    \{b'_j\},\{s'_j\}=\Theta(x'_{u};\theta)
\end{equation}
$\{b'_j\},\{s'_j\}$ are the sets of boxes and class-wise scores of predictions on $x'_{U}$. $j$ denotes the $j$-th prediction.

As shown in Fig.~\ref{fig:detail}, in the first stage we extract individual information from images by consistency-based metric $M$ between reference and matching predictions. According to the rank of $M$, we form an initial selected pool which is slightly over the budget. In the second stage, we evaluate the mutual information between samples in the initial pool and labeled pool and decide the final selected pool to meet the annotation budget.

\vspace{0.7em}
\noindent \textbf{3.3.1 Consistency-based Individual Information (Stage 1)}

\vspace{0.7em}
\noindent In order to calculate the consistency of predictions, first we need to match a corresponding prediction (including the \textbf{corresponding box} $b'_k$ and \textbf{corresponding score} $s'_k$ in $\{b'_j\},\{s'_j\}$) to each reference prediction $b_k$. 
We choose the corresponding box $b'_k$ that has the maximum Intersection over Union ($IoU$) with $b_k$. $s'_k$ is the score of $b'_k$. The matching process $b_k \leftrightarrow b'_k$ can be formulated as:
\begin{equation}
    b'_k =	\argmax_{b'_j \in \{b'_j\}}IoU(b'_j,b_k).
\end{equation}

The next step is to compute the consistency between reference prediction and corresponding prediction. For box regression, we directly use $IoU$ which can clearly indicate the matching degree of two boxes. To measure the distance between two class-wise probabilities, Jensen-Shannon ($JS$) divergence and Kullback-Leibler (KL) divergence are popular metrics. We specifically take advantage of $JS$ since it has clear upper and lower bounds, allowing us to quantify it in conjunction with $IoU$. Besides divergence of possibilities, we also adopt the maximum confidence as a weight factor to emphasize the prediction with high confidence. This is because a high-confidence prediction has a greater impact on performance (in case multiple predictions correspond to the same ground truth, only the prediction with the highest confidence is regarded as true positive, while others will be regraded as false positives).  Finally, we reverse $JS$ to $1-JS$ to keep the same trend as $IoU$. The consistency of the $k$-th prediction of an image can be computed as the sum of consistencies of boxes $C^{b}_k$ and scores $C^{s}_k$:  
\begin{equation}
\label{eq:mk}
m_k = C^{b}_k + C^{s}_k \\
\end{equation}
\noindent where
\begin{equation}
\label{eq:Ck}
\begin{gathered}
C^{b}_k=IoU(b_k,b'_k) \\
C^{s}_k= \underbrace{\frac{1}{2}[\max_{\varphi_{n} \in s_k} (\varphi_{n})+ \max_{\varphi_{n}' \in s'_k} (\varphi_{n}')]}_{\text{weight factor}}(1-JS(s_k||s'_k))
\end{gathered}
\end{equation}

\begin{figure}[t]
    \centering
    \begin{subfigure}{.3\linewidth}
        \includegraphics[height=1.8cm]{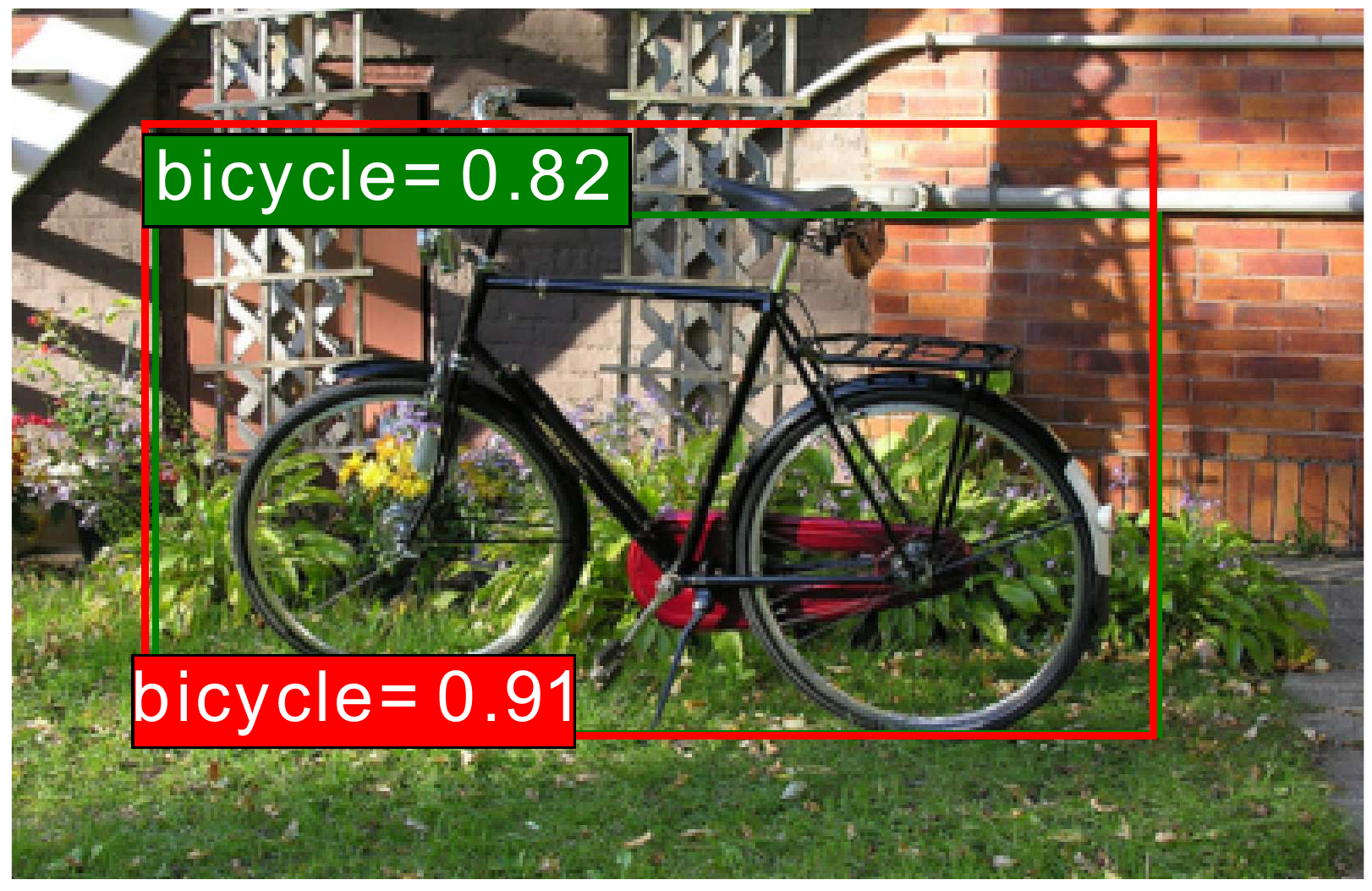}\vspace{-2pt}
        \caption{}
        \label{fig:case_a}
    \end{subfigure}\hspace{0.3in}
    \begin{subfigure}{.3\linewidth}
        \includegraphics[height=1.8cm]{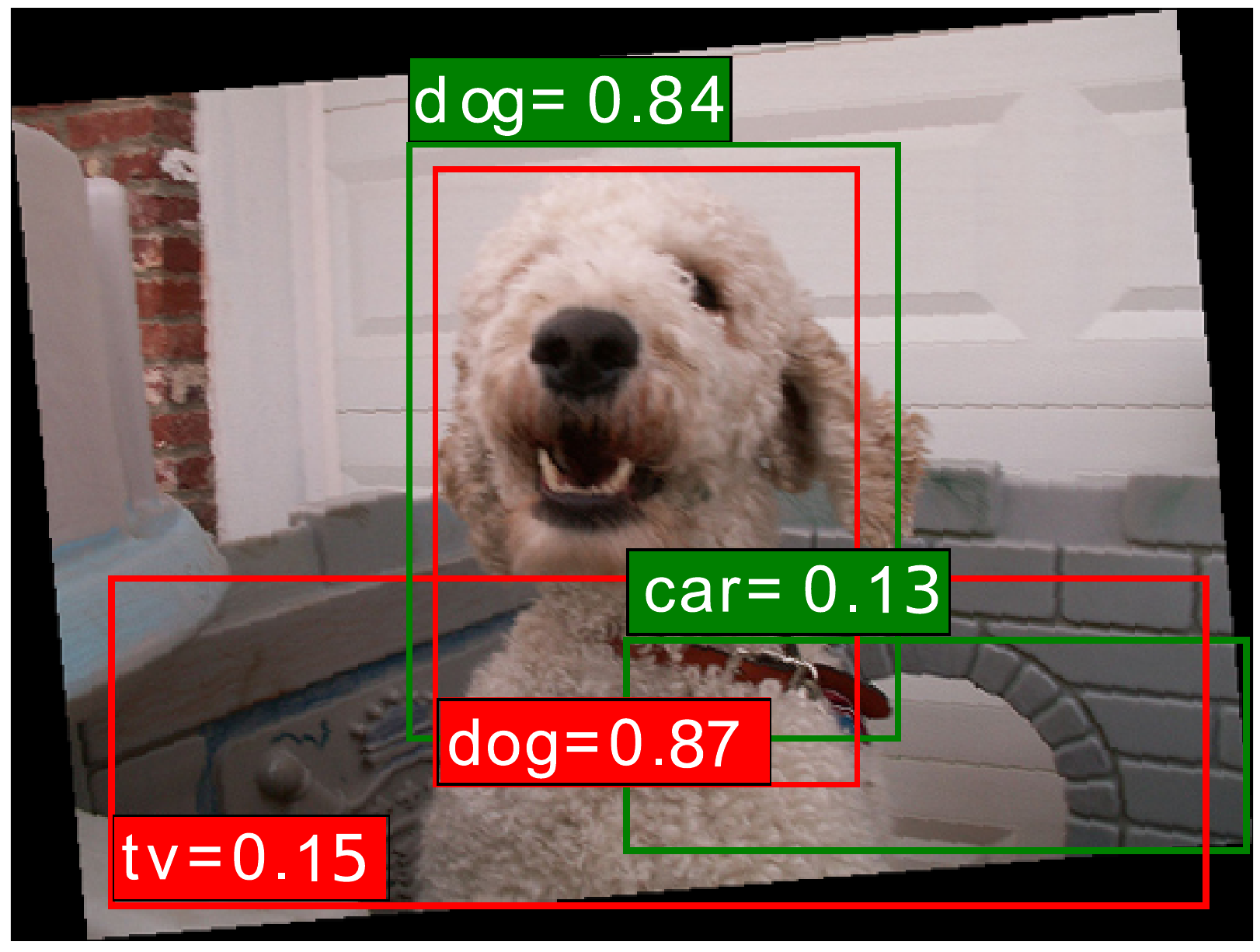}\vspace{-2pt}
        \caption{}
        \label{fig:case_b}
    \end{subfigure}
    

    \begin{subfigure}{.3\linewidth}
        \includegraphics[height=1.8cm]{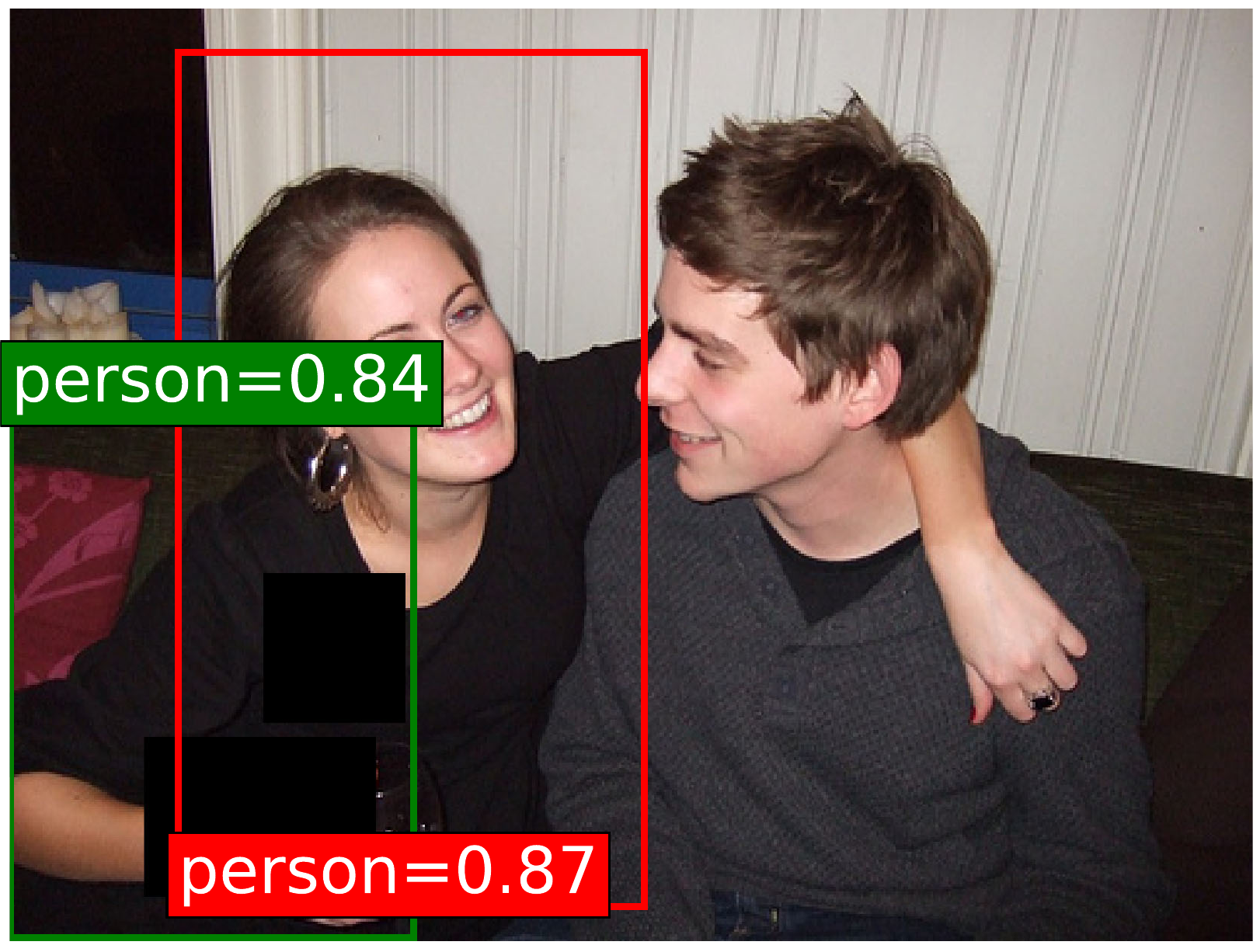}\vspace{-2pt}
        \caption{}
        \label{fig:case_c}
    \end{subfigure}
    \begin{subfigure}{.3\linewidth}
        \includegraphics[height=1.8cm]{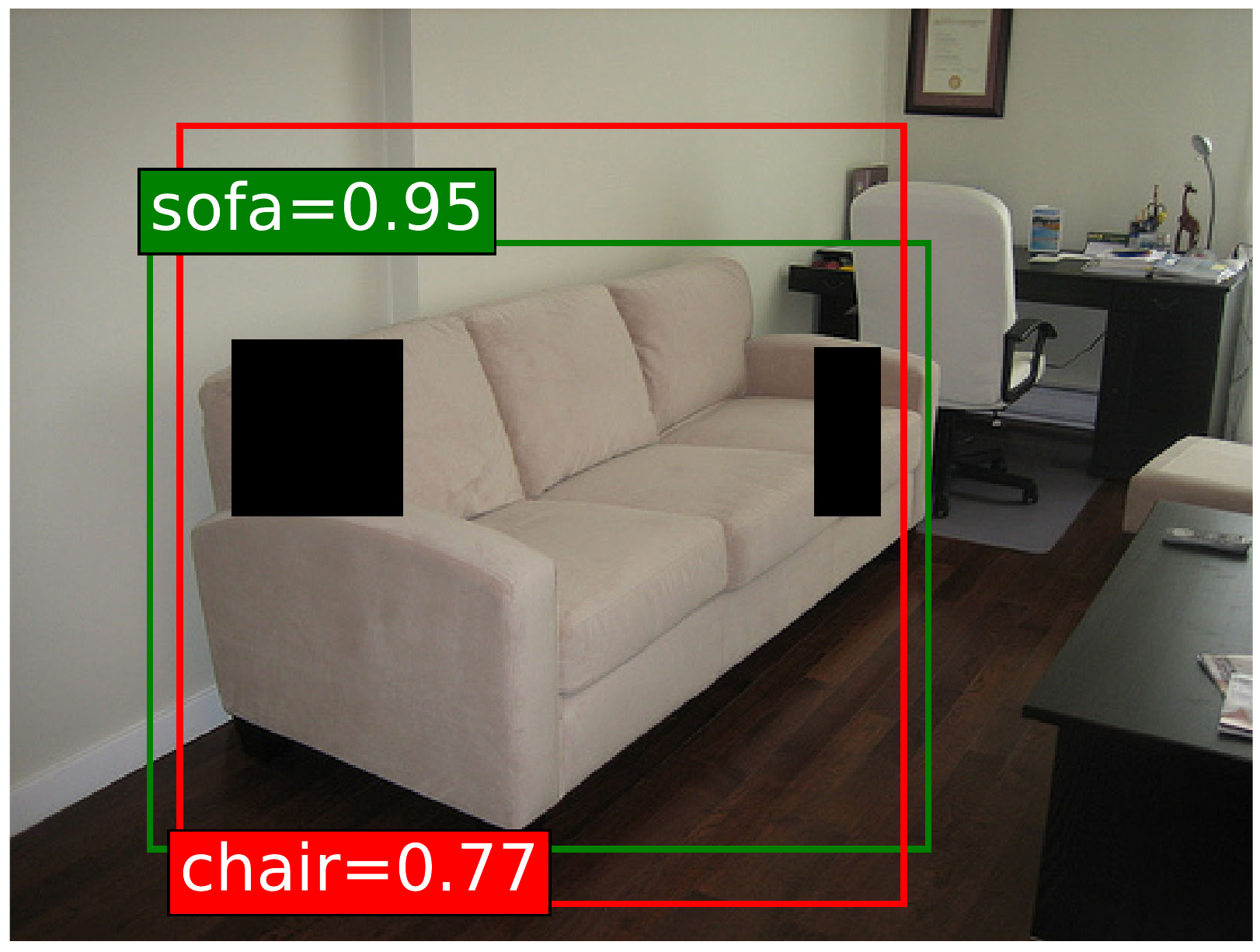}\vspace{-2pt}
        \caption{}
        \label{fig:case_d}
    \end{subfigure}
    \begin{subfigure}{.3\linewidth}
        \includegraphics[height=1.8cm]{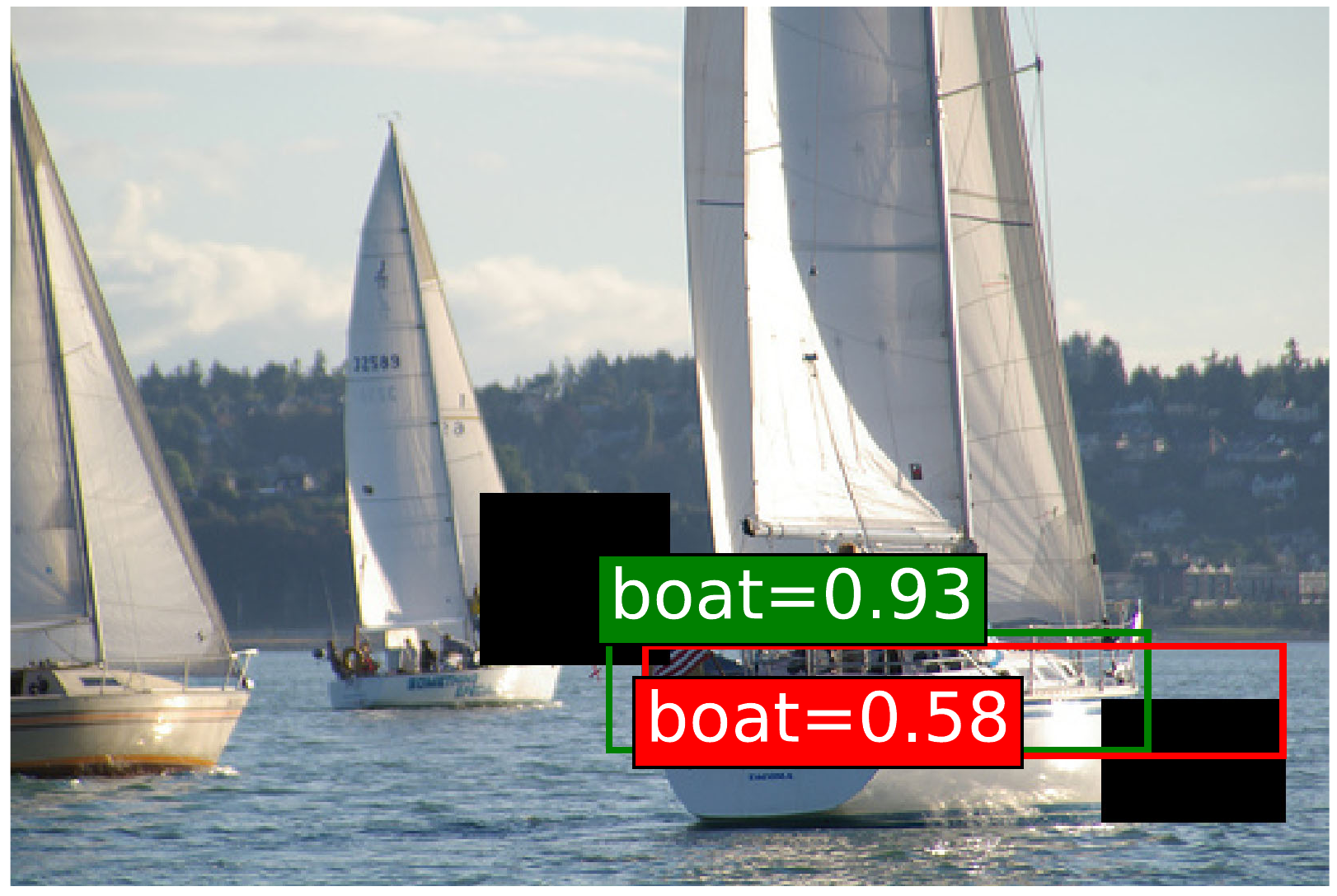}\vspace{-2pt}
        \caption{}
        \label{fig:case_e}
    \end{subfigure}
    \caption{Case analysis. Red boxes are reference predictions and green boxes are their corresponding predictions. (a)(b) are uninformative cases while (c)(d)(e) are informative cases.}
\label{fig:case}
\end{figure}

Obviously, $m_k \in [0,2]$. Ideally, a smaller consistency $m_k$ indicates a more unstable prediction, which also implies more informative the patch is. \textit{However, directly define $m_k$ as the metric of information does not work well in practice.} The most important question is when $m_k$ is on the lower bound, is the prediction the most informative one? The answer is not necessarily. For example, as shown in Fig.~\ref{fig:case_b}, for the paired predictions  \textit{\textcolor{red}{tv}} (i.e. reference prediction) and \textit{\textcolor{ForestGreen}{car}} (i.e. corresponding prediction), $C^{b}_k$ and $C^{s}_k$ are both small and $m_k$ is close to its lower bound. However, due to the bad matching and low confidence, this prediction is very unlikely to be the main result. The reason is that the detector may give another accurate prediction of the \textit{dog}, which is also shown in Fig.~\ref{fig:case_b}. In this case, we can observe that the predictions for the main object \textit{dog} are actually very stable and accurate, manifesting an uninformative sample to the detector. But if simply using $m_k$ (the lower the better), this sample is falsely considered as an informative one. In other words, the smallest $m_k$ does not necessarily represent the most informative patch in practice due to the instability or randomness of prediction. 



On the other hand, when $m_k$ is close to the upper bound (e.g. Fig.~\ref{fig:case_a}), the detector can handle the augmentation well and give a high-confidence prediction which is likely to be correct due to the high matching degree. Such samples are not informative because the detector can deal with the augmentations well. Based on these observations and analyses, we speculate that $m_k$ of an informative prediction 
should have two properties:
(1) Keeping a certain distance from the lower bound, which means the paired predictions have relative high matching degree and high confidence. If the prediction is wrong, this patch is probably informative, because the prediction is likely the main result of the object (the detector will not give other accurate predictions of the object like the case in Fig.~\ref{fig:case_b}). And this is based on the fact that there cannot be multiple predictions with high confidence in the same area at the same time, according to Softmax and non-maximum suppression (NMS). (2) Being far away from the upper bound, which means the matching degree is worse than when $m_k$ is on the upper bound (such as Fig.~\ref{fig:case_a}). This indicates that the detector cannot cope with common augmentations on the image, and this prediction is likely to be inaccurate (such as Figs.~\ref{fig:case_c}, \ref{fig:case_d} and \ref{fig:case_e}).


To quantify this, the consistency-based metric of an image is defined as
\begin{equation}
\label{equ:M}
    M(x_u;\mathcal{A},\Theta)=\mathbb{E}_\mathcal{A}[\min_{k} \vert m_k-\beta \vert],
\end{equation}
where $\beta$ is the base point to represent $m_k$ of the most informative patch. Based on the above analysis, we search the optimal $\beta$ heuristically: starting from the midpoint of the upper and lower bounds of $m_k$, the optimal $\beta$ can be found with several grid searching steps (we only use 5 steps).
The optimal value of $\beta$ searched by this procedure is effective for all datasets and detectors. The reason we adopt minimum value over an image instead of mean value is that we focus on finding the most informative local regions instead of the whole image. Finally, we compute the expectation of $M$ over multiple augmentations to improve reliability.

\begin{figure}[ht]
\begin{center}
 \includegraphics[width=0.95\linewidth]{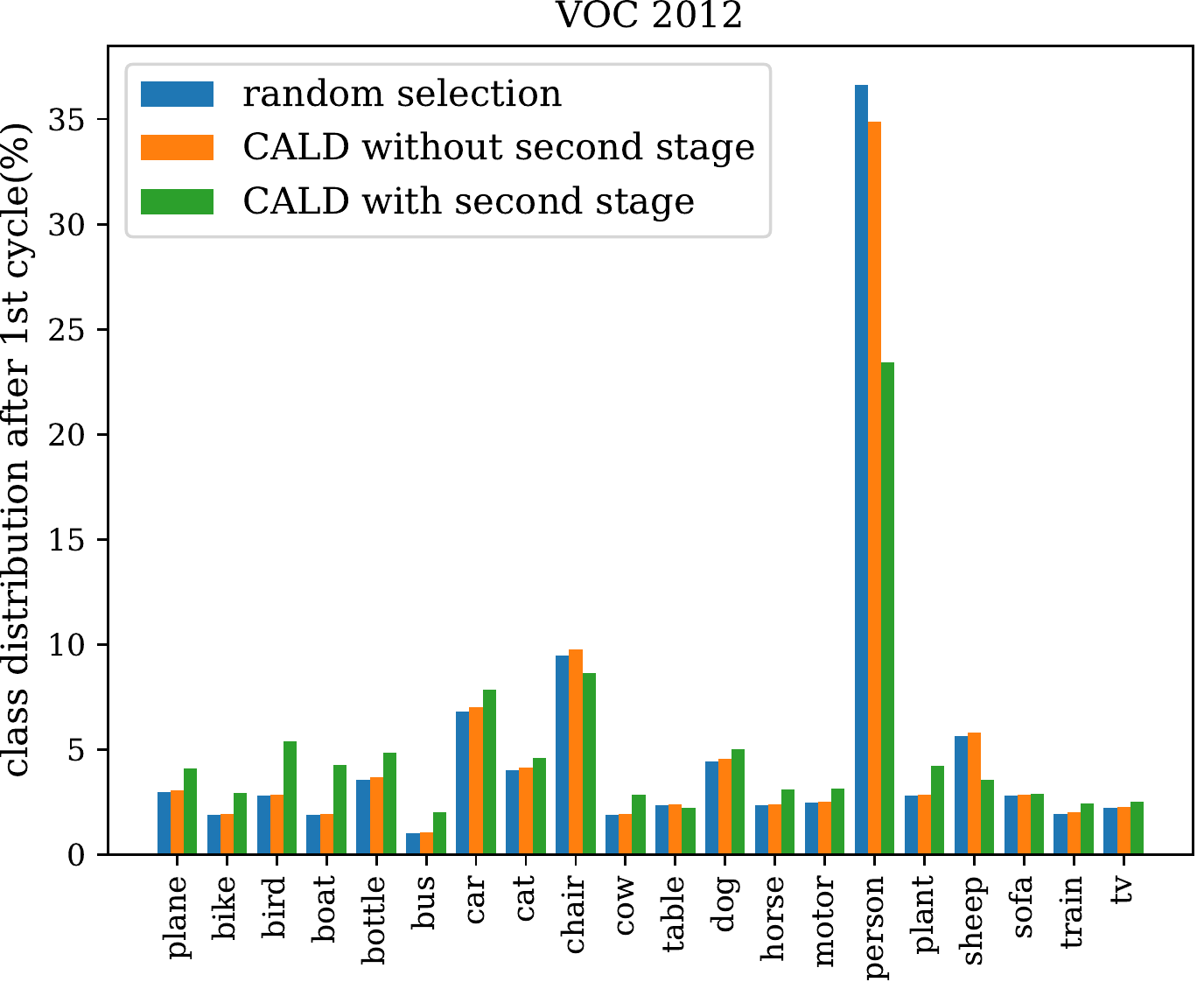}
\end{center}
\vspace{-13pt}
   \caption{Class distributions of the labeled pool (after the first active learning cycle) with different selection strategies. 
   }
\label{fig:distribution}
\end{figure}

\vspace{0.7em}
\noindent \textbf{3.3.2 Mutual Information (Stage 2)}

\vspace{0.7em}
\noindent We find that the class distribution of the labeled pool is unbalanced after random selection. As shown in Fig.~\ref{fig:distribution}, the height differences between the blue bars are very large. And this issue cannot be solved by only using individual information (orange bars) for sample selection since there are multiple objects in detection images.

We propose an inter-image metric to alleviate the issue. The idea is to compare the class distribution of each sample in the initial selected pool and that of the whole labeled pool, then select samples with large distances (i.e. having different class distributions from the labeled pool) to form the final selected pool. We use $JS$ divergence to evaluate the distance between two class distributions (i.e. mutual information).
The procedures of computing the mutual information are outlined in Algorithm~\ref{alg:mutual}. For the labeled pool, we sum all ground truth ($Y_L$) to represent the class distribution which is computed as
\begin{equation}
\begin{gathered}
\Delta_{L}(Y_{L})=Softmax([\delta_{1},\delta_{2},\cdots,\textcolor{blue}{\delta_{m}},\cdots]^\mathrm{T}),\\
\hspace{-1.55cm} \textcolor{blue}{\delta_{m}}=\sum_{y_L \in Y_L}\mathcal{I}(y_L=m).
\end{gathered}
\end{equation}
$m$ denotes the $m$-th category in the dataset and $\mathcal{I}$ is the indicator function. For an unlabeled image $x_U$ in $X_I$, we only count the highest confidence of predictions in each class due to high certainty. If we follow the notations in Sec.~\ref{sec:method:problem}, denoting the class-wise classification prediction of original and augmented image by $s_k$ and $s'_j$ in which $\varphi_m$ ($\varphi_m'$) is the score of $m$-th class, the process can be formulated as
\begin{equation}
\small
\begin{gathered}
\Delta_{U}(x_{U})=Softmax([\delta_{1},\delta_{2},\cdots,\textcolor{blue}{\delta_{m}},\cdots]^\mathrm{T})\\
\textcolor{blue}{\delta_{m}}=\max_{s_{k} \in \{s_{k}\}} \{\varphi_m | \varphi_m \in s_k\}+\max_{s'_{j} \in \{s'_{j}\}} \{\varphi_m' |\varphi_m' \in s'_j\}\\
\end{gathered}
\end{equation}

Return to Fig.~\ref{fig:distribution}, we can observe that after selecting by mutual information (green bars), in general, the proportions of the majority categories (such as \textit{person}) have dropped while the proportions of minority categories (such as \textit{bus} and \textit{bike}) have risen, alleviating the unbalanced distribution.
\renewcommand{\algorithmicrequire}{\textbf{Input:}}
\renewcommand{\algorithmicensure}{\textbf{Functions:}}
\begin{algorithm}[t]
\small
\caption{Selection by mutual information in each cycle}
\label{alg:mutual}
\begin{algorithmic}
\REQUIRE {Initial selected pool $X_I$, ground truth of labeled pool $Y_L$, total budget $B$, budget per cycle $B/C$}
\ENSURE {Distribution function of labeled pool $\Delta_{L}(Y_L)$ and single unlabeled image $\Delta_{U}(x_U)$ }

\STATE $X_F \leftarrow \{\}$
\WHILE{size($X_F$) \textless $B/C$}
\STATE $\textcolor{blue}{f} = \argmax_{\textcolor{red}{x_{U} \in X_I}}[JS(\Delta_{U}(x_U)||\Delta_{L}(Y_L)]$
\STATE $X_F = X_F \cup \{X_I[\textcolor{blue}{f}]\}$
\STATE $X_I = X_I - \{X_I[\textcolor{blue}{f}])\}$
\ENDWHILE
\RETURN $X_F$
\end{algorithmic}
\label{a1}
\end{algorithm}

\begin{figure*}[t]
\begin{center}
 \includegraphics[width=1.0\linewidth]{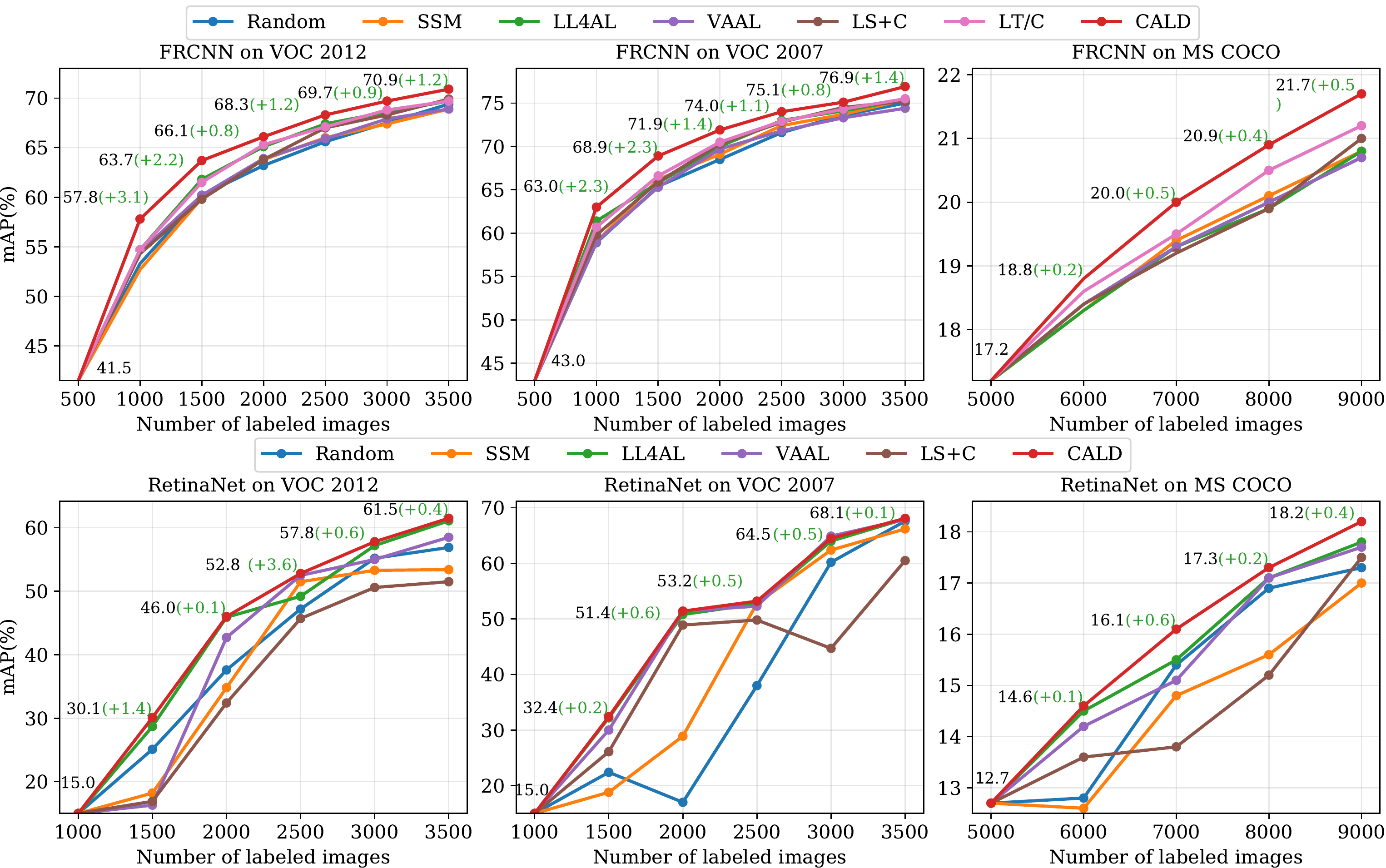}
\end{center}
\vspace{-11pt}
   \caption{Comparison with SOTA task-agnostic and detection-specific active learning methods (LT/C only applies to two-stage detectors). CALD surpasses all methods comprehensively on three datasets and two detectors. The numbers marked on the points of CALD denote performance and its improvement over the second-best method. In the first row, the second-best methods are all LT/C~\cite{kao2018localization} while the second-best methods are all LL4AL~\cite{yoo2019learning} in the second row.}
\label{fig:results}
\end{figure*}

\section{Experiments}
\label{sec:experiment}
\noindent \textbf{Datasets.}
To validate the effectiveness of CALD, we conduct extensive experiments on three benchmarks for object detection:  MS COCO~\cite{lin2014microsoft}, Pascal VOC 2007 and VOC 2012~\cite{everingham2010pascal}. On VOC 2012 and COCO, we use training set for training and validation set for testing. On VOC 2007, we use trainval set for training and test set for testing. On VOC, we set 500 labeled images as random initialization and 500 as budget per cycle. For RetinaNet, we set 1000 as initialization since 500 images are too few to train a robust model for RetinaNet. On COCO, we set 5000 as initialization and 1000 as budget per cycle by following~\cite{kao2018localization}. Detectors are evaluated with mean Average Precision (mAP) at IoU $=0.5$ on VOC and with average mAP from IoU $=0.5$ to IoU $=0.95$ on COCO, which are standard protocols for these datasets.

\noindent \textbf{Detectors.}
We employ the popular two-stage detector Faster R-CNN (FRCNN)~\cite{ren2016faster} and single-stage detector RetinaNet~\cite{lin2017focal}, both with Resnet50~\cite{he2016deep} and FPN~\cite{lin2017feature}, as task models. The implementation of the two models follows the default settings of Torchvision. In each cycle we train the models for 20 epochs. \textit{The numbers reported in the results are averages of 3 trials for each method and detector.}

\subsection{Comparison with State of the Art }
We compare the proposed CALD with random selection (Random), three detection-specific active learning methods (SSM~\cite{wang2018towards}, LS+C and LT/C~\cite{kao2018localization}) and two task-agnostic active learning methods (VAAL~\cite{sinha2019variational} and LL4AL~\cite{yoo2019learning}), which represent the state-of-the-art (SOTA). As shown in Fig.~\ref{fig:results}, CALD outperforms the SOTA methods on all three datasets with both FRCNN and RetinaNet detectors.

On VOC 2012 and VOC 2007 with FRCNN, the improvements of CALD over random selection and the second-best method are significant. Specifically, in terms of mAP, CALD is 8.4\% and 7.0\% higher than random selection, and 5.7\% and 3.8\% higher than the second-best method LT/C in the first cycle on VOC 2012 and 2007, respectively. This demonstrates the effectiveness of CALD by following the three guidelines: unifying the metric of box regression and classification, focusing on local regions and promoting a balanced class distribution. Also the improvements manifest a consistent trend: in the first cycle, the improvements are the largest and gradually decrease in subsequent cycles (3.1 to 1.2 and 2.3 to 1.4). The reason is that as the number of available unlabeled samples gradually decreases, samples collected by all methods tend to be the same.
\begin{figure}[t]
\begin{center}
 \includegraphics[width=0.99\linewidth]{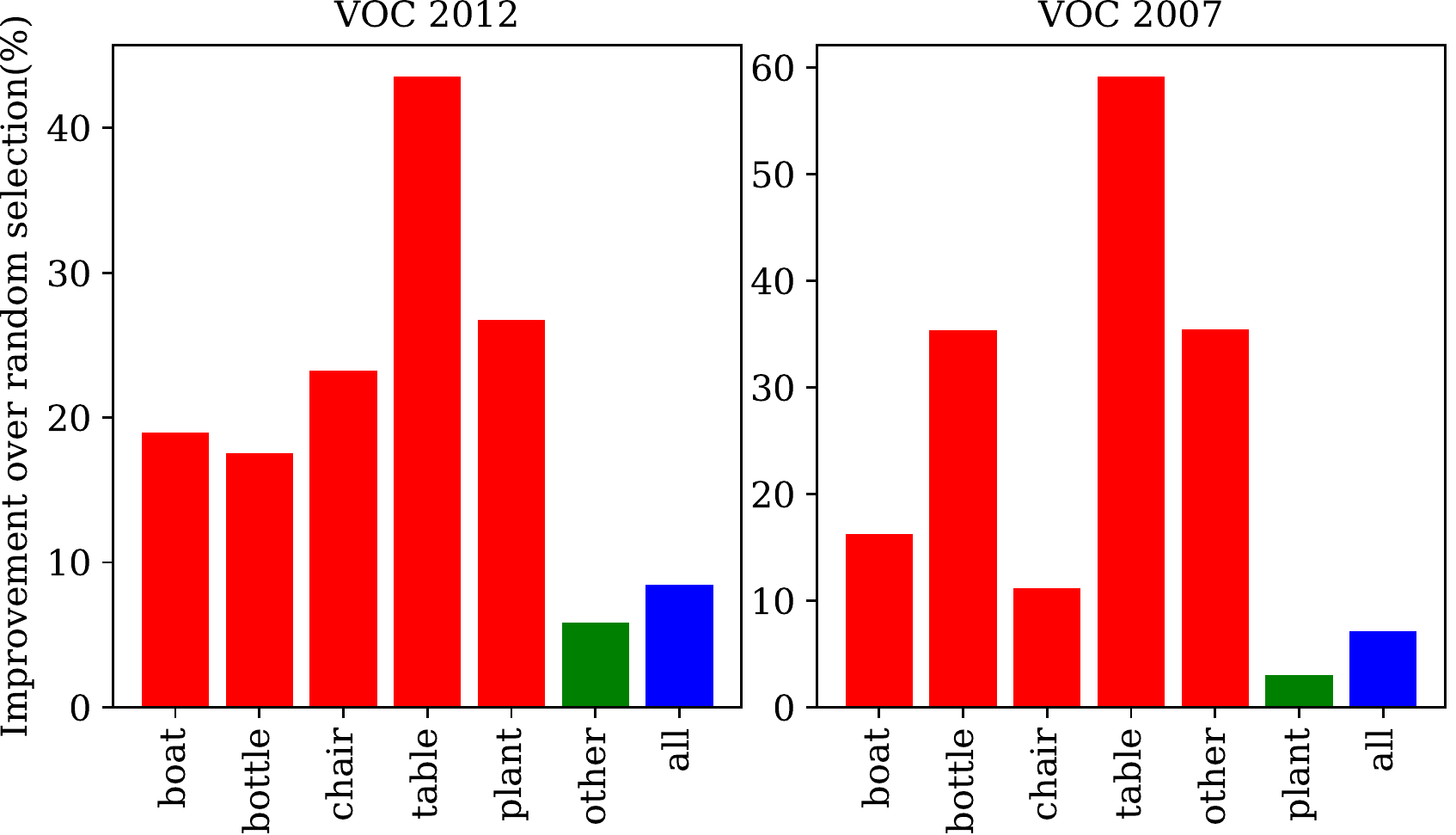}
\end{center}
\vspace{-10pt}
   \caption{The improvement in difficult classes (red bars) over random selection in the first cycle on VOC. Green and  blue bars are improvements for non-difficult classes (others) and all classes.}
\label{fig:difficult}
\end{figure}
We observe that the task-agnostic active learning methods LL4AL and VAAL perform bad (5.4\% and 5.7\% lower than CALD in the first cycle on VOC 2012) on two-stage detector. FRCNN first extracts region proposals and then adopts fine-grained predictions on local patches. Such complexity widens the gap between detection and classification, leading to worse performance of classification-based methods.
LT/C, which is specifically designed for two-stage detectors, performs second only to CALD (2.5\% lower than CALD on average on VOC 2012). However, its shortcomings are also obvious. First, it cannot be widely used in other detectors (such as one-stage detectors) while CALD can be generalized to any detector. Secondly, it cannot process the classification information finely, since FRCNN dose not give class-wise scores in the first stage. Although LS+C has considered box regression, it does not combine boxes and classification to get a comprehensive metric, so it does not perform well in practice.
\begin{figure*}[t]
    \centering
    \begin{subfigure}{0.34\textwidth}
        \includegraphics[width=1\textwidth]{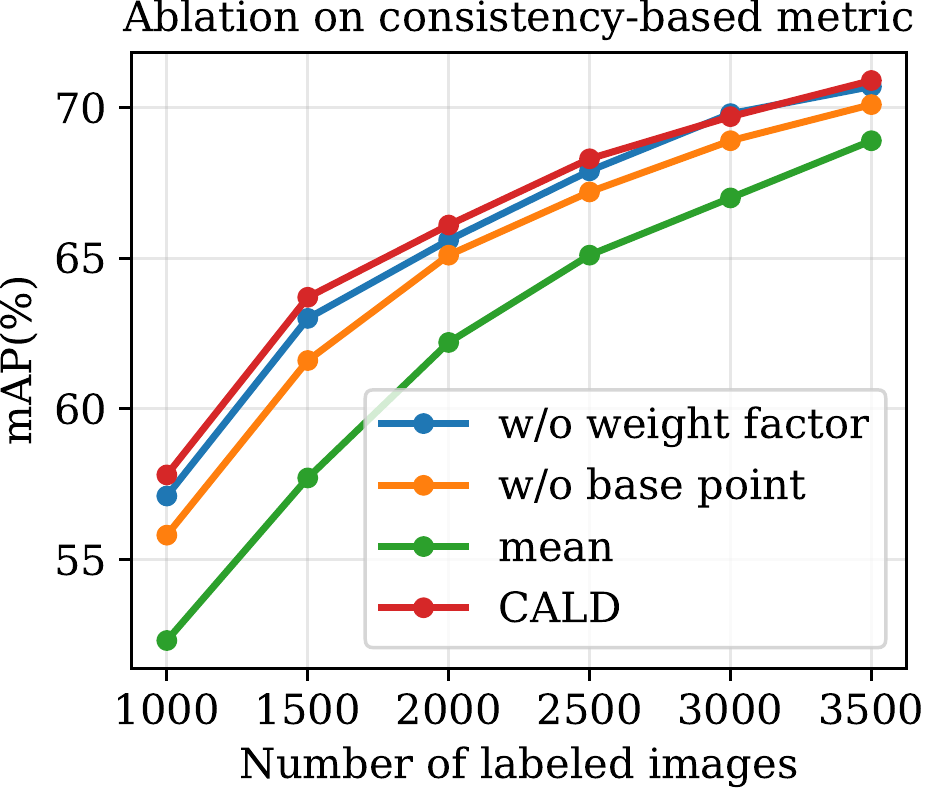}
        \vspace{-10pt}
        \caption{}
        \label{fig:ablation_a}
    \end{subfigure}
    \begin{subfigure}{0.32\textwidth}
        \includegraphics[width=1\textwidth]{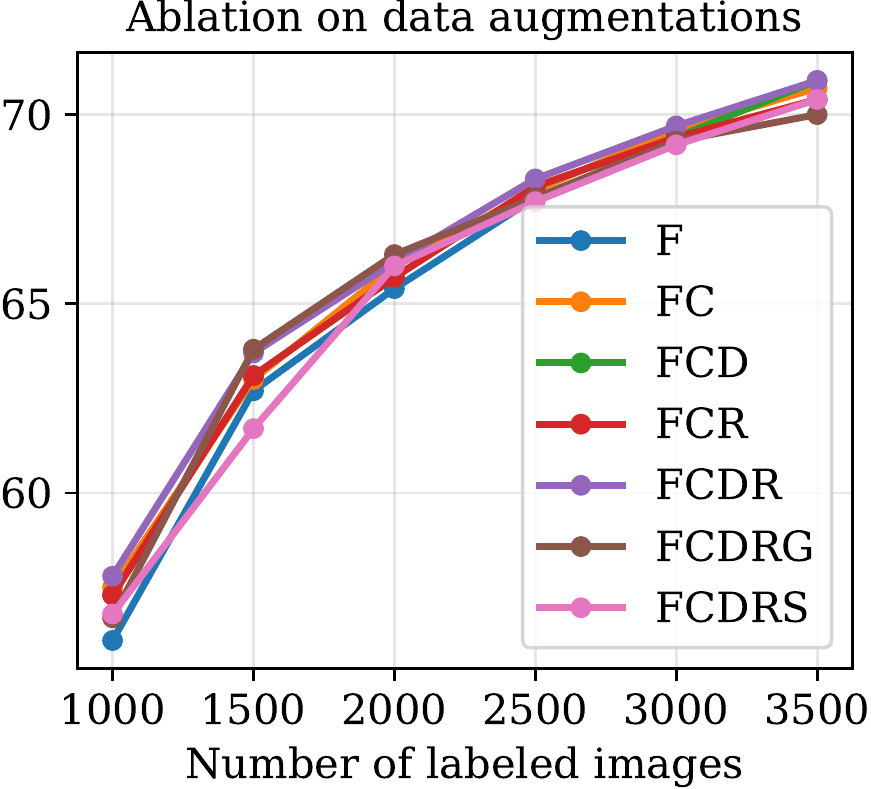}
        \vspace{-10pt}
        \caption{}
        \label{fig:ablation_b}
    \end{subfigure}
    \begin{subfigure}{0.32\textwidth}
        \includegraphics[width=1\textwidth]{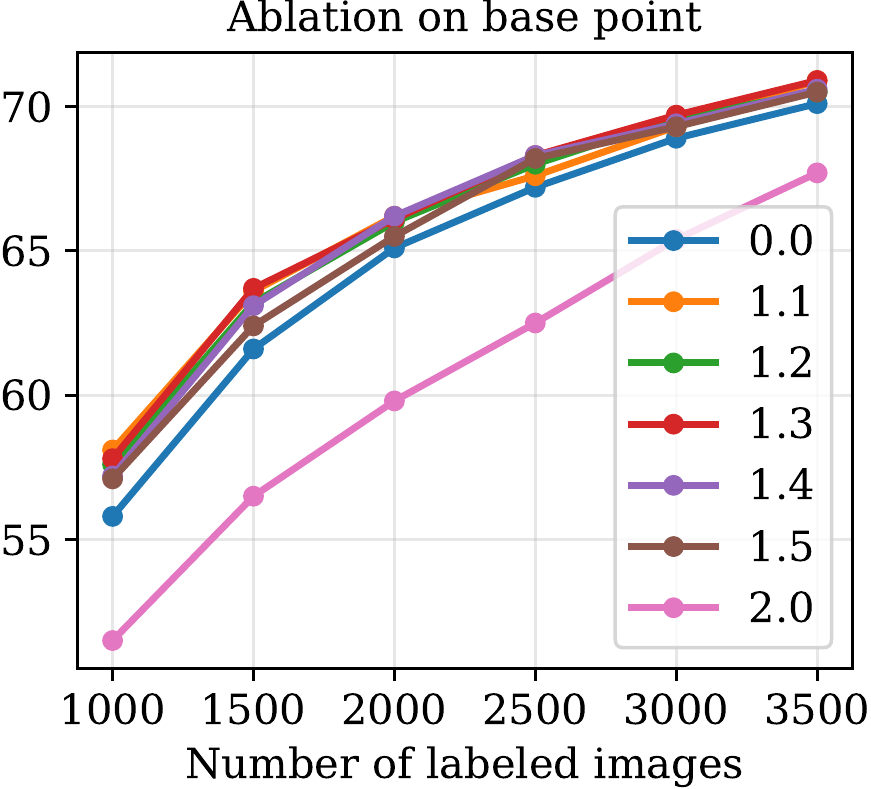}
        \vspace{-10pt}
        \caption{}
        \label{fig:ablation_c}
    \end{subfigure}
    \vspace{-5pt}
    \caption{Ablation studies on strategies of consistency-based metric, data augmentations, and base point $\beta$.}
\label{fig:ablation}
\end{figure*}

As for RetinaNet, the improvement of CALD is also the most significant: 11.8\% higher than random selection on average on VOC 2012. Compared with FRCNN, the performance of LL4AL using RetinaNet is slightly closer to CALD (2.5\% lower than CLAD on average on VOC 2012). The reason is that RetinaNet has a simpler architecture which directly gets predictions from global features, therefore the global information has a greater impact on the final results than that of FRCNN. However, classification-based methods still cannot take box regression into account. 


We also note that CALD yields
more improvements in difficult categories. For categories with AP lower than 40\% in random selection, we treat them as difficult categories. For the difficult categories (red bars) in Fig.~\ref{fig:difficult}, we notice that the improvements are larger than other classes.


\subsection{Ablation Study}
We conduct ablation studies on VOC 2012 with FRCNN. 

\noindent \textbf{Consistency-based metric.}
To validate the consistency-based metric $M$ is reasonable, we conduct ablation study on different strategies. Firstly, in Eq.~\ref{equ:M}, we use the minimum $|m_k-\beta|$ of an image. For ablation study, we investigate the performance of using mean $|m_k-\beta|$ for an image, i.e.
\begin{equation}
\label{equ:M_mean}
    M(x_u;\mathcal{A},\Theta)=\mathbb{E}_\mathcal{A}[\mathbb{E}_{k} \vert m_k-\beta \vert],
\end{equation}
which represents the average global information of the image. As shown in Fig. \ref{fig:ablation_a}, the performance (curve of ``mean") drops significantly because detectors focus more on local regions rather than global information. Secondly, we remove the base point $\beta$, i.e. let $\beta$ equal to 0. 
The performance drops since the smallest $m_k$ does not necessarily represent the most informative patch as we discussed in \textbf{Sec. 3.3.1}
with a case study of Fig.~\ref{fig:case_b}.
Finally, if we remove the weight factor in $m_k$ (see Eq. \ref{eq:Ck}), which means we treat all predictions with different confidences equally, the performance also drops slightly. Because prediction with high confidence has a greater impact on performance.

\noindent \textbf{Data augmentations.}
We compute the detection consistency based on common data augmentations in our method. For simplicity, we use a single uppercase letter to denote one type of augmentation. ``F'' for horizontal flip, ``C'' for cutout, ``D'' for downsize, ``R'' for rotation, ``G'' for Gaussian noise, and ``S'' for salt and pepper noise. The combination of letters means we get $M$ by averaging the results of these augmentations. As shown in Fig.~\ref{fig:ablation_b}, CALD works well with standard augmentations, and does not rely on specific augmentations.
The proper combination of augmentations can make the performance more stable. We adopt ``FCDR'' in CALD.

\noindent \textbf{Base point $\beta$.}
Base point is the parameter $\beta$ of consistency-based metric in Eq.~\ref{equ:M}, which denotes the value of $m_k$ of the most informative prediction. From the plots in Fig.~\ref{fig:ablation_c}, $\beta$ gets the optimal value around 1.3. Thus $\beta=1.3$ is used for all datasets and detectors in our experiments. When the value of $\beta$ goes from the optimal point to the lower bound (0.0) of $m_k$, the performance of CALD decreases slowly. It can be explained that when $m_k$ is closer to the lower bound, the predictions become unstable which are not necessarily informative. On the contrary, if $\beta$ is excessively closer to the upper bound, the performance drop quickly. This is because $m_k$ close to the upper bound denotes uninformative predictions (cases like Fig.~\ref{fig:case_a}). When $m_k$ reaches the upper bound (2.0), CALD selects the least informative samples (performance of detector is even worse than random selection), which also indicates that CALD can clearly distinguish whether the sample is informative or not.

\begin{table}
\small

\begin{center}
\renewcommand{\arraystretch}{0.8}
\begin{tabular}{c|c|c}
\hline
Expansion ratio & mAP of 1st cycle  & mAP of 2nd cycle \\
\hline
0\% &  56.9  & 62.8   \\
10\% &  57.4  & 63.3   \\
20\% &  \textbf{57.8}  & \textbf{63.7}   \\
30\% &  57.3  & 63.5   \\
40\% &  57.1  & 63.7  \\
\hline
\end{tabular}
\end{center}
\vspace{-10pt}
\caption{Ablation on the expansion ratio for $X_I$.}
\label{tab:extended}
\end{table}

\noindent \textbf{Expansion ratio for $X_I$}. As stated in Sec.~\ref{sec:consist}, we form the initial selected pool $X_I$ in the first stage by selecting more samples than the budget, so that we can further filter those samples in the second stage to meet the budget for each cycle. Then one question emerges: \textit{how many more samples to use?} Assume the budget of each cycle is 500 images, selecting 600 images for $X_I$ in the first stage means a 20\% expansion ratio. 
We investigate different expansion ratios and report the results in Table~\ref{tab:extended}. Note that 0\% in this table indicates our method reduces to \textbf{only have the first stage}. We reach two conclusions. (1) Based on the results of 0\%, 10\% and 20\%, there is a clear advantage of leveraging mutual information for sample selection in the second stage. (2) 20\% additional budget for $X_I$ yields the best performance, leading to an mAP improvement of 0.9 in both cycles (56.9 \textit{vs.} 57.8; 62.8 \textit{vs.} 63.7). However, keep expanding the budget in the first stage would also cause performance drop (e.g. 30\% ratio). This is because more informative samples may be removed by mutual information in the second stage in order to cut back to the fixed budget. Therefore, the experimental results reveal the importance of both individual and mutual information for sample selection.

\section{Conclusion}
\label{sec:conclusion}
This paper introduces a consistency-based active learning method for object detection, namely CALD. In order to select the most informative samples, it 
leverages a consistency-based metric to consider the information of box regression and classification simultaneously, which is ignored by previous methods. 
In addition to sample individual information, CALD also uses mutual information to refine sample selection to encourage a balanced class distribution.
Extensive experiments show that CALD with different detectors achieves state-of-the-art performance on several object detection benchmarks under active learning settings. 

{\small
\bibliographystyle{ieee_fullname}
\bibliography{egbib}
}

\end{document}